\documentclass[sigconf,utf8,utf8x]{acmart}
\usepackage{graphicx}
\usepackage{amsmath}

\usepackage{amssymb}
\usepackage{booktabs}
\usepackage{microtype}
\usepackage{balance}
\usepackage{multirow}
\usepackage{caption}
\usepackage{algorithm}
\usepackage[noend]{algorithmic}
\usepackage{mathtools}
\usepackage{bm}
\usepackage{CJKutf8}
\usepackage{subfig}
\AtBeginDocument{%
  }

\copyrightyear{2023}
\acmYear{2023}
\setcopyright{acmlicensed}\acmConference[MM '23]{Proceedings of the 31st ACM International Conference on Multimedia}{October 29-November 3, 2023}{Ottawa, ON, Canada}
\acmBooktitle{Proceedings of the 31st ACM International Conference on Multimedia (MM '23), October 29-November 3, 2023, Ottawa, ON, Canada}
\acmPrice{15.00}
\acmDOI{10.1145/3581783.3612265}
\acmISBN{979-8-4007-0108-5/23/10}
\begin{document}

\title{Resolve Domain Conflicts for Generalizable Remote Physiological Measurement}

\author{Weiyu Sun}
\email{weiyusun@smail.nju.edu.cn}
\orcid{0000-0002-3948-0705}
\affiliation{%
  \institution{Nanjing University}
  % \streetaddress{P.O. Box 1212}
  \city{Nanjing}
  \state{Jiangsu}
  \country{China}
}
% \authornote{Corresponding authors}

\author{Xinyu Zhang}
\email{xinyuzhang@smail.nju.edu.cn}
\orcid{0009-0003-1894-3541}
\affiliation{%
  \institution{Nanjing University}
  % \streetaddress{P.O. Box 1212}
  \city{Nanjing}
  \state{Jiangsu}
  \country{China}
}

\author{Hao Lu}
\email{hlu585@connect.ust.hk}
\orcid{0000-0002-2241-6598}
\affiliation{%
  \institution{The Hong Kong University of Science and Technology}
  % \streetaddress{P.O. Box 1212}
  \city{Guangzhou}
  \state{Guangdong}
  \country{China}
}

\author{Ying Chen}
\email{yingchen@nju.edu.cn}
\orcid{0000-0001-9086-2832}
\authornote{Corresponding authors}
\affiliation{%
  \institution{Nanjing University}
  % \streetaddress{P.O. Box 1212}
  \city{Nanjing}
  \state{Jiangsu}
  \country{China}
}

\author{Yun Ge}
\email{geyun@nju.edu.cn}
\orcid{0000-0003-1788-3295}
\affiliation{%
  \institution{Nanjing University}
  % \streetaddress{P.O. Box 1212}
  \city{Nanjing}
  \state{Jiangsu}
  \country{China}
}

\author{Xiaolin Huang}
\email{xlhuang@nju.edu.cn}
\orcid{0000-0002-3382-5346}
\affiliation{%
  \institution{Nanjing University}
  % \streetaddress{P.O. Box 1212}
  \city{Nanjing}
  \state{Jiangsu}
  \country{China}
}

\author{Jie Yuan}
\email{yuanjie@nju.edu.cn}
\orcid{0000-0003-0597-7657}
\affiliation{%
  \institution{Nanjing University}
  % \streetaddress{P.O. Box 1212}
  \city{Nanjing}
  \state{Jiangsu}
  \country{China}
}

\author{Yingcong Chen}
\email{yingcong.ian.chen@gmail.com}
\orcid{0000-0002-9565-8205}
\affiliation{%
  \institution{The Hong Kong University of Science and Technology}
  % \streetaddress{P.O. Box 1212}
  \city{Guangzhou}
  \state{Guangdong}
  \country{China}
}

\renewcommand{\shortauthors}{Weiyu Sun et al.}

% \begin{abstract}
% Remote photoplethysmography (rPPG) technology has become increasingly popular due to its non-invasive monitoring of various physiological indicators, making it widely applicable in multimedia interaction, healthcare, and emotion analysis. Existing rPPG methods utilize multiple datasets for training to enhance the generalizability of models. However, they often overlook the underlying conflict issues across different datasets, such as (1) label conflict resulting from different phase delays between physiological signal labels and face videos at the instance level, and (2) attribute conflict stemming from distribution shifts caused by head movements, illumination changes, skin types, etc. To address this, we introduce the DOmain-HArmonious framework (DOHA). Specifically, we first propose a harmonious phase strategy to eliminate uncertain phase delays and preserve the temporal variation of physiological signals. Next, we design a harmonious hyperplane optimization that reduces irrelevant attribute shifts and encourages the model's optimization towards a global solution that fits more valid scenarios. Our experiments demonstrate that DOHA significantly improves the performance of existing methods under multiple protocols. We plan to release the source codes after acceptance.
% \vspace{-1mm}
% \end{abstract}

\begin{abstract}
Remote photoplethysmography (rPPG) technology has become increasingly popular due to its non-invasive monitoring of various physiological indicators, making it widely applicable in multimedia interaction, healthcare, and emotion analysis. Existing rPPG methods utilize multiple datasets for training to enhance the generalizability of models. However, they often overlook the underlying conflict issues across different datasets, such as (1) label conflict resulting from different phase delays between physiological signal labels and face videos at the instance level, and (2) attribute conflict stemming from distribution shifts caused by head movements, illumination changes, skin types, etc. To address this, we introduce the DOmain-HArmonious framework (DOHA). Specifically, we first propose a harmonious phase strategy to eliminate uncertain phase delays and preserve the temporal variation of physiological signals. Next, we design a harmonious hyperplane optimization that reduces irrelevant attribute shifts and encourages the model's optimization towards a global solution that fits more valid scenarios. Our experiments demonstrate that DOHA significantly improves the performance of existing methods under multiple protocols. Our code is available at \url{https://github.com/SWY666/rPPG-DOHA}.
\vspace{-1mm}
\end{abstract}

\vspace{-5mm}
\ccsdesc[500]{Applied computing~Health informatics}

\vspace{-5mm}
\keywords{physiological signal estimation, rPPG, multimedia application}
\vspace{-3mm}

% \received{}
% \received[revised]{}
% \received[accepted]{}

\maketitle

\begin{figure}[ht]
    \centering
    \includegraphics[width=8.4cm]{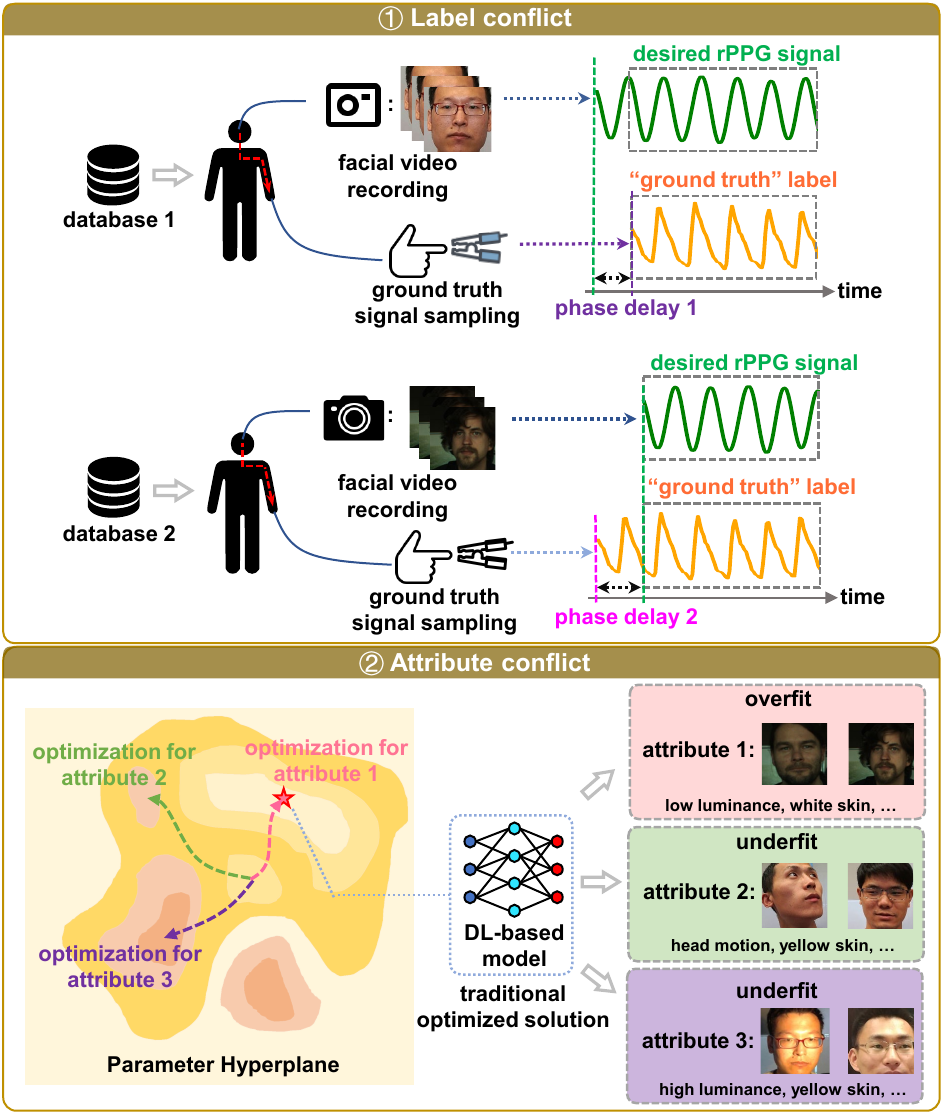}
    \vspace{-3mm}
    \caption{Contradictory factors exist in rPPG model training. \begin{CJK}{UTF8}{gbsn}①\end{CJK}: Label conflicts caused by different pulse transit times (red dashed line) and sampling equipment, which result in the chaotic phase delays between desired rPPG signals and ``ground truth'' signals among different databases. \begin{CJK}{UTF8}{gbsn}②\end{CJK}: Irrelevant attribute conflicts raised by diverse external noise (e.g., illumination, head motion). Such diversity and the predominance of the intensity over the rPPG signal make the optimization difficult to cover all involved domains.}
    \label{Two_conflictions}
    \vspace{-6.5mm}
\end{figure}

\vspace{-1mm}
\section{Introduction} \label{Introduction}

With wide application in multimedia \cite{Psy_app, phys_apply7_ACM, NIRP, mcduff2023camera}, remote photoplethysmography (rPPG) can estimate various physiological indicators from facial video and further apply to fields such as human-computer interaction \cite{HCI1, HCI2}, emotional computing \cite{emotion1, emotion2} and security authentication \cite{spoof_rppg1, spoof_rppg2, spoof_rppg3}. Unlike traditional physiological measurement techniques like photoplethysmography (PPG), which require physical contact \cite{wearable}, rPPG offers a more convenient and cost-effective solution. By analyzing chrominance variations \cite{Euler-Enlarge} in facial videos, rPPG can extract the underlying physiological periodic signal known as the rPPG signal. This can be achieved effortlessly, even through a basic webcam. The rPPG signal carries information about the cardiac activity, which can be derived using established practices used for other signal types such as blood volume pulse (BVP) signals \cite{STVEN, PhysFormer}. The advantage of rPPG lies in its non-contact nature and the ability to obtain physiological information remotely. Despite the potential advantages of rPPG, the fragility of the rPPG signal poses challenges for traditional algorithm-based rPPG methods \cite{POS, GREEN, CHROM} when faced with demanding tasks, such as videos that suffer from illumination variation or head movement.

Recently, deep learning (DL) networks have emerged as effective tools for assisting rPPG signal extraction. With the accumulation of existing works \cite{RhythmNet-prelude,zhangsenle, RhythmNet, CVD, Dual-GAN, Deepphys,CAN,PhysNet,PhysFormer,ETA_rPPG,HR-CNN,Meta-rppg,VitaMon, siamese_rPPG}, DL-based rPPG methods have demonstrated significantly better performance than traditional methods\cite{GREEN,CHROM,POS,PCA,ICA} over various intra-dataset \cite{VIPL-HR, PURE, UBFC} conditions. But in turn their performance against unseen scenarios is less impressive \cite{NEST}. Therefore, for better practical performance, currently multiple datasets are used for network optimization \cite{NEST} to improve the generalization of these deep learning models.

However, there are many obstacles preventing DL rPPG models from benefiting via multiple dataset training. These obstacles mainly derive from the domain conflicts between different databases. As shown in Fig. \ref{Two_conflictions}, we conclude these conflicts into two categories: label conflict and attribute conflict. The label conflict mainly refers to the phase delay differentials raised by different pulse transit times \cite{PTT, VitaMon} of subjects and different ``ground truth’’ label sampling equipment. These phase delay differentials reduce the label stability, which can lead to pool learning efficiency for the DL rPPG model. The attribute conflict, on the other hand, is raised by the vulnerability of rPPG signal \cite{GREEN, POS} against external attributes (e.g., illumination, head motion, skin type) in facial videos. The rPPG signal is feeble (pixel level) \cite{POS}, which makes the DL model easily dominated by the aforementioned external attributes (or domain-specific features). As a result, DL-based models tend to learn the relationship between different irrelevant attributes and different label delays, which leads to the failure to learn a generalized physiological representation.

In this paper, we propose the DOmain-HArmonious framework (DOHA), which addresses both label conflict and attribute conflict for rPPG tasks. Specifically, we design a harmonious phase strategy (DOHA-HPS) that transfers the temporal physiological information into a self-similar representation (termed as self-similarity physiological map) to alleviate label conflicts. This strategy preserves the temporal information of physiological signals and eliminates the negative effects of elusive phase differentials. Then, we propose a harmonious hyperplane optimization (DOHA-HHO) to reduce domain shift overfitting. DOHA-HHO first identifies those out-of-distribution instances based on their gradient norms and then restricts their impacts, termed as global gradient harmony (DOHA-GGH). Next, we perform instance-wise gradient surgery (namely instance-wise gradient harmony, DOHA-IGH) to alleviate conflicting components for the remaining instances in the batch, thus letting the model focus on the common feature (i.e. rPPG signal) among instances. This novel optimization strategy can not only alleviate model attributes overfitting at the instance level but also seek domain-shared features at the global level. Our contributions are summarized below:

$\bullet$ We define the domain conflict problem of the rPPG task, which limits the generalization ability of the model when using multiple datasets simultaneously.

$\bullet$ We propose a harmonious phase strategy, which can eliminate the negative effects of uncertain phase delays with maximally preserving the temporal information of physiological signals.

$\bullet$ We propose a harmonious hyperplane optimization strategy, which can alleviate irrelevant attributes overfitting at both instance and global levels.

$\bullet$ We provide extensive experimental results to demonstrate the effectiveness of DOHA. 

\vspace{-2mm}
\section{Related Work} \label{Related Work}
\subsection{Remote Photoplethysmography Technique}
Remote photoplethysmography (rPPG) is one of the techniques of remote physiological measurement. It estimates HR and other health indexes (e.g., heart rate variability (HRV)) via analyzing the chrominance change (termed as rPPG signal) from the facial video. Early traditional rPPG methods \cite{GREEN, CHROM, ICA, PCA, POS} rely on the pre-defined algorithm based on physiological prior knowledge. Recently, the competence of deep learning contributes to various remarkable DL-based rPPG methods \cite{STVEN, PhysNet, CAN, Deepphys, PhysFormer, RhythmNet, CVD, Dual-GAN, HR-CNN}. These DL-based rPPG methods can be divided into two categories: end-to-end methods \cite{STVEN, PhysNet, CAN, Deepphys, PhysFormer, HR-CNN} that take in original facial videos, and non-end-to-end methods \cite{RhythmNet, CVD, Dual-GAN, NEST} that first transform facial videos into specialized spatial-temporal (ST) maps before analysis. Despite their promising results, label conflict remains a challenge for these methods, and existing solutions are either costly (e.g., manual calibration \cite{PhysNet}) or limited in their usage scenarios (e.g., requiring manual preprocessing \cite{RhythmNet}).

\begin{figure*}[ht]
    \centering
    \includegraphics[width=16.5cm]{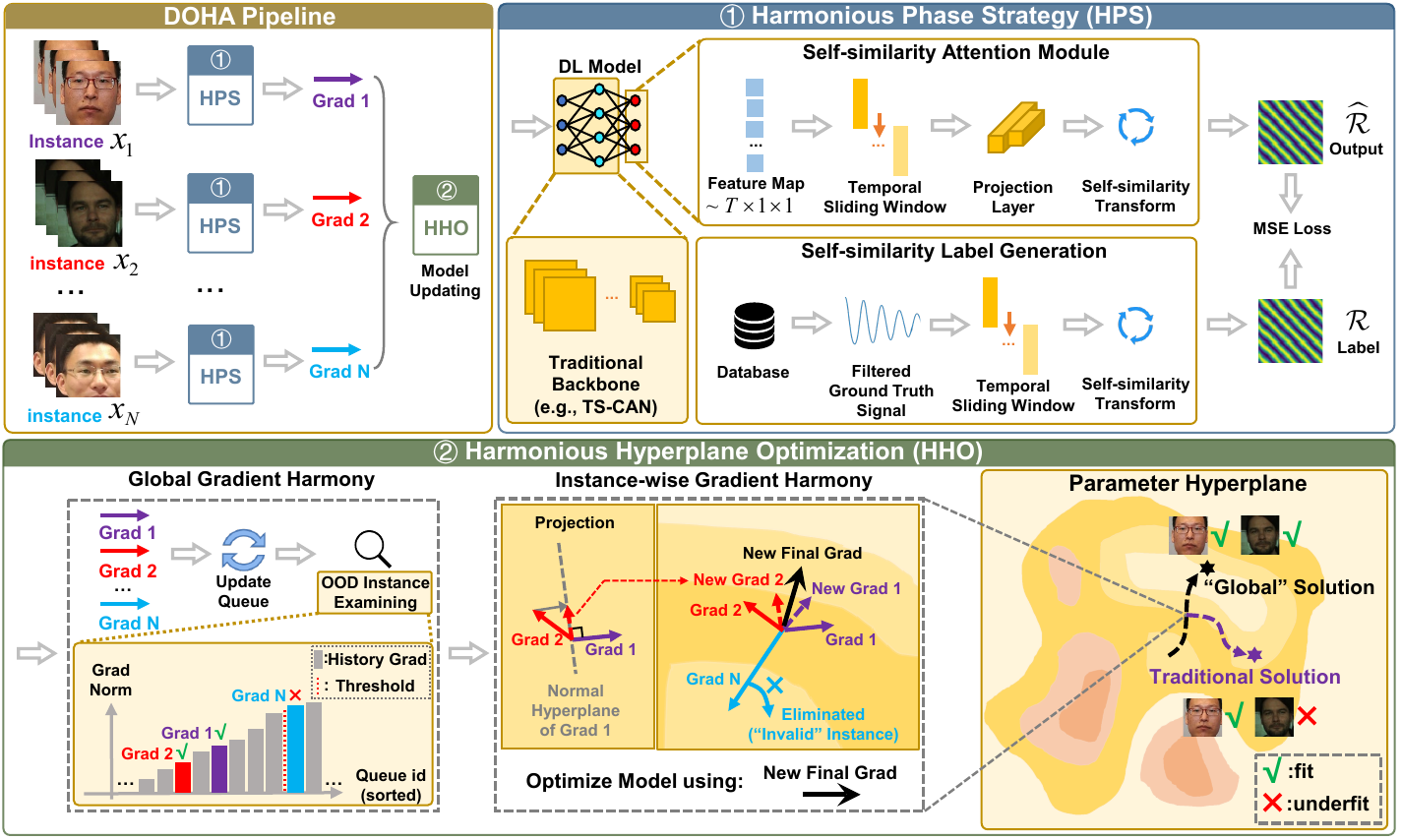}
    \vspace{-3mm}
    \caption{An overview of our proposed method. \begin{CJK}{UTF8}{gbsn}①\end{CJK} exhibits our harmonious phase strategy on ground truth signal and network output, respectively. $\hat{\mathcal{R}}$ and $\mathcal{R}$ present typical network output and label in the form of self-similarity physiological map. \begin{CJK}{UTF8}{gbsn}②\end{CJK}: a brief demonstration of Harmonious Hyperplane Optimization. Compared with the traditional optimization rule, it eliminates potential invalid instances (instance $X_N$) and then de-conflicts remaining instances on the parameter hyperplane, so that the optimization can cover more attributes.}
    \label{Main-work-flow}
    \vspace{-4.5mm}
\end{figure*}

\vspace{-2mm}
\subsection{Domain Generalization}
The domain generalization (DG) technique has shown promise in addressing the domain conflict problem discussed in Section \ref{Introduction}. DG techniques aim to enhance the model's ability to handle domain shifts, thereby improving its performance on unseen scenarios \cite{DS_survey, Domain_ACM1, Domain_ACM2, Domain_ACM3, AD, NCDG, DomainGS}. In the context of rPPG, various DG-related approaches have been explored \cite{NEST, chung2022domain}, including data augmentation methods \cite{sun2022contrast}, meta-learning \cite{Meta-rppg}, and representation learning \cite{NEST}. However, these techniques often overlook or underestimate the negative effect of domain conflicts, resulting in insufficient improvement. The gradient surgery (GS) technique is a special form of the DG method that addresses potential conflicts during model training. It operates on backward gradients in the parameter hyperplane \cite{DomainGS, GradientSurgery}, enabling effective optimization towards desired directions. This technique has been successfully applied in various training tasks to address conflicts between different tasks \cite{GradientSurgery} or different domains \cite{DomainGS}. However, the rPPG databases present unique complexities, as conflicts can arise at the instance level, and the difficulty of different instances can vary greatly and be elusive. This phenomenon poses challenges to the effectiveness of traditional gradient surgery \cite{DomainGS}, which relies on accurate domain labels (e.g., dataset classes). To address these challenges, we propose a specialized GS-based method called Harmonious Hyperplane Optimization for rPPG. This method offers a fine-grained and easy-to-implement approach to mitigate conflicts in the rPPG task.

\vspace{-3mm}
\section{Methodology}

\vspace{-1mm}
\subsection{Overall Framework}
\vspace{-1mm}
In this work, we propose the DOmain-HArmonious framework (DOHA) to alleviate the aforementioned conflicts, as illustrated in Fig. \ref{Main-work-flow}. First, for each instance in the batch, we collect the corresponding backward gradient via the harmonious phase strategy (DOHA-HPS, Fig. \ref{Main-work-flow} \begin{CJK}{UTF8}{gbsn}①\end{CJK}). DOHA-HPS transforms the ground truth signal and model output into the self-similarity space, represented by $\mathcal{R}$ and $\hat{\mathcal{R}}$ respectively, to neutralize label conflicts. Next, we perform harmonious hyperplane optimization (DOHA-HHO, Fig. \ref{Main-work-flow} \begin{CJK}{UTF8}{gbsn}②\end{CJK}) on the collected gradients. It consists of two parts: (1) Global gradient harmony (DOHA-GGH), which identifies and excludes the disharmonious instances that are not suitable for the overall optimization process. (2) Instance-wise gradient harmony (DOHA-IGH), which mitigates the conflicting portions among the remaining instances, preventing the model from overfitting on irrelevant attributes and underfitting on others.

\vspace{-3mm}
\subsection{Harmonious Phase Strategy} \label{Invariant Feature Learning}

\textbf{Motivation}. Phase delay differentials in rPPG databases are common, unavoidable, and elusive. These differentials arise from two main factors: (1) variations in the ``ground truth" signal sampling equipment, where each device introduces different delays due to circuit design, and (2) pulse transit time (PTT) \cite{VitaMon, PTT}, which represents the time it takes for the physiological signal to transit through the subject's body (indicated by the red arrow in Fig. \ref{Two_conflictions}). The former factor leads to conflicts in labels across datasets, while the latter can cause conflicts at the instance level. These conflicts cannot be eliminated due to the inherent physiological nature and limitations of the techniques used. Although these conflicts can be observed in instances that can be analyzed by traditional rPPG methods (e.g., GREEN \footnote{Traditional methods estimate the rPPG signal directly from the input video, instead of the sampling equipment, thus providing a reliable reference for delay detection. Readers can refer to the Appendix. \ref{Limitation of signal calibration} for more details.}), existing rPPG methods \cite{GREEN, CHROM, POS} have poor resistance against external noise (such as head motion), resulting in only a small fraction of instances being suitable for such analysis. Consequently, many valuable instances would be discarded, leading to both labor-intensive efforts and reduced data utilization. Therefore, there is a need for a cost-effective and efficient approach to address label conflicts.

\begin{figure}
    \centering
    \includegraphics[width=8cm]{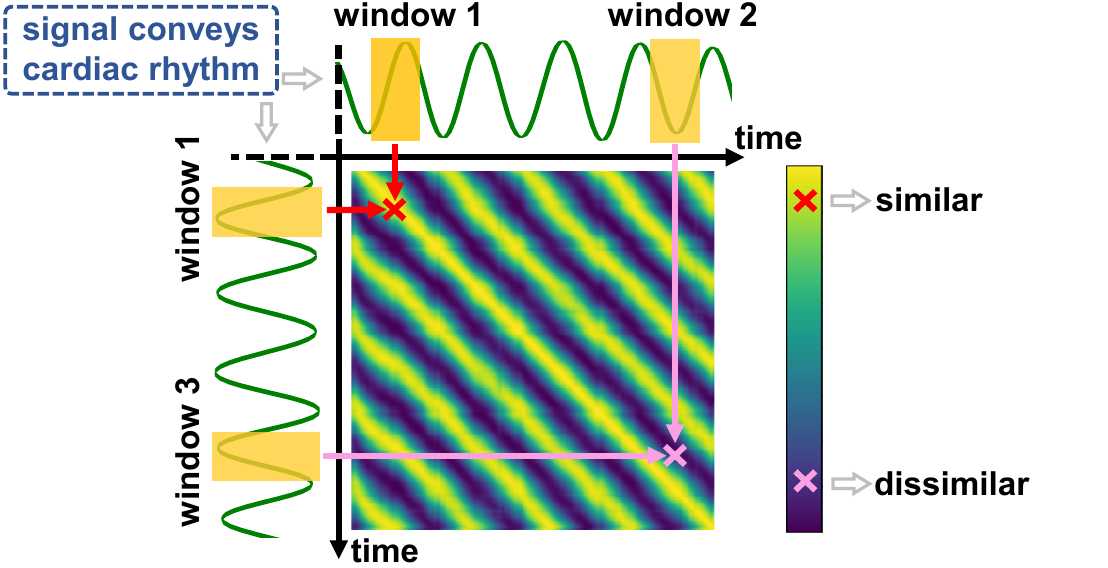}
    \vspace{-3mm}
    \caption{A typical self-similarity physiological map, where a higher value (bright) indicates that the temporal messages between two windows are similar, and vice versa. The ``signal'' can be a ground truth wave (e.g., BVP signal), or the predicted signal of the rPPG model.}
    \label{SSP}
    \vspace{-5mm}
\end{figure}

\noindent \textbf{Self-similarity physiological map}. To this end, we propose the Harmonious Phase Strategy (DOHA-HPS) that provides an informative and stable label representation. DOHA-HPS leverages the concept of temporal self-similarity to mitigate the effects of phase delay while preserving important temporal physiological information. Specifically, we first divide the ``ground truth'' signal into different time windows using a sliding window and calculate the similarity (e.g., cosine similarity) among windows to form a matrix $\mathcal{R}$, as shown in Fig. \ref{SSP} \footnote{Detailed generation for $\mathcal{R}$ and $\hat{\mathcal{R}}$ can refer to Appendix \ref{Algorithm for SSP map generation} and algorithm \ref{SCS Map}.}. Each element $\mathcal{R}_{ij}$ indicates the similarity between the $i_{th}$ and $j_{th}$ temporal windows. As each element in $\mathcal{R}$ depends only on the temporal distance between two windows and the cardiac period, the impact of delay differential is thus minimized.

\noindent \textbf{Self-similarity attention module}. Similarly, we can generate an output $\hat{\mathcal{R}}$ at the end of the rPPG model, such as the BVP signal estimation head \cite{NEST}, as shown in Fig. \ref{Main-work-flow} \begin{CJK}{UTF8}{gbsn}①\end{CJK}. And to better analyze temporal information, we insert a projection layer (e.g., linear layer) between temporal windows and self-similarity transform. As a result, both $\mathcal{R}$ and $\hat{\mathcal{R}}$ contain aligned temporal information and are derived from the subject's cardiac activity. Therefore, we can define loss functions based on the similarity between $\mathcal{R}$ and $\hat{\mathcal{R}}$ to optimize the model. For sure, many potential loss function designs can be explored, and here we provide a typical loss item (i.e. Mean Squared Error (MSE) loss) for optimization, as shown in eq. (\ref{MSE_loss}).

\begin{equation}
    % \vspace{-2mm}
    \label{MSE_loss}
    \mathcal{L}_m = \frac{1}{N^2}\sum_{i=0}^{N-1}\sum_{j=0}^{N-1}(\mathcal{R}_{ij}-\hat{\mathcal{R}}_{ij})^{2}
\end{equation}

\noindent where $N$ is the size of self-similarity physiological map $\mathcal{R}$ and $\hat{\mathcal{R}}$.

\noindent \textbf{Cardiac information inversion}. In the network output $\hat{\mathcal{R}}_{N \times N}$, the elements belonging to the set $\{\hat{\mathcal{R}}_{ij}|i-j=t_m\}$ describe the relationship between two windows with a temporal distance of $t_m$ ($t_m \in [0, N-1]$). By collecting element sets that belong to different $t_m$ in $\hat{\mathcal{R}}$, we thus obtain an autocorrelation sequence $Seq$ of the desired rPPG signal, from which we can extract physiological messages such as the heart rate ($HR$). The detailed algorithm is depicted in Algorithm \ref{cardiac extraction}, where $\mathcal{F}(\cdot)$ is a band-pass filter with a range of $[0.7, 3.5]Hz$, and $\mathcal{D}(\cdot)$ analyzes peak distance \footnote{Such as the function ``signal.find\_peaks'' in ``scipy'' package.} to determine the frequency (related with heart rate) of $Seq$. Noteworthy, more robust and effective methods for extracting information from $\hat{\mathcal{R}}$ can be explored, such as non-end-to-end methods \cite{RhythmNet, zhangsenle} or two-dimensional spectrum analysis, which we leave for future research.

\begin{algorithm}
% \vspace{-1.5mm}
	\caption{Cardiac Information Inversion} 
	\label{cardiac extraction} 
	\begin{algorithmic}[1]
            \STATE {\bfseries Input:} network output $\hat{\mathcal{R}}_{N \times N}$, output size $N$, band-pass filter $\mathcal{F}(\cdot)$, peak detection algorithm $\mathcal{D}(\cdot)$.
            \STATE initialize autocorrelation sequence $Seq$: $\{\}$
            \STATE $Seq[t_m] \leftarrow \{\hat{\mathcal{R}}_{ij}|i-j=t_m\}, \forall t_m \in [0, N-1]$
            \STATE $Seq[t_m] \leftarrow mean(Seq[t_m])$
            \STATE $HR = \mathcal{D}(\mathcal{F}(Seq))$
            \STATE {\bfseries Output:} heart rate $HR$
	\end{algorithmic}
% \vspace{-1mm}
\end{algorithm}
% \vspace{-4mm}

\vspace{-3mm}
\subsection{Harmonious Hyperplane Optimization}
\label{Harmonious rPPG DL Training Methodology}
\begin{figure}
    \centering
    \includegraphics[width=6.5cm]{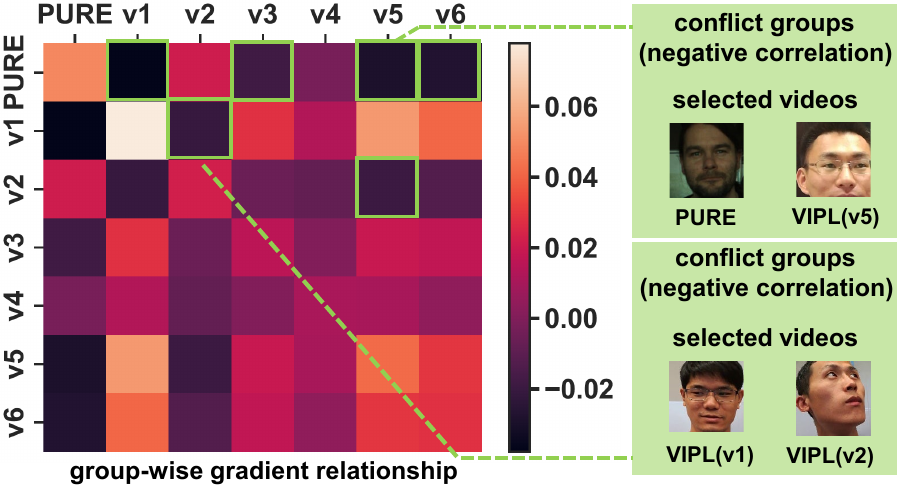}
    \vspace{-3mm}
    \caption{The gradient of baseline \cite{CAN} has conflicts among both intra- and inter-dataset. For example, the average gradient of VIPL (subset v1-v6) \cite{VIPL-HR} and PURE \cite{PURE} prominently present the negative cosine similarity.}
    \label{Conflict_domain_pre}
    \vspace{-4mm}
\end{figure}

\textbf{Motivation}. Compared to external factors (e.g., illumination variation and head motion), the rPPG signal is very feeble, leading the learning of the rPPG model vulnerable to irrelevant attribute conflicts. To verify this, we examine the relationship (i.e. cosine similarity) of the backward gradients among instances from PURE and VIPL (task v1-v6 \footnote{The detailed information of task v1-v6 of VIPL database can refer to the Appendix \ref{Details on involved databases}.}) databases on a DL-based model (TS-CAN \cite{CAN}) pre-trained using these databases, as shown in Fig. \ref{Conflict_domain_pre}. According to the illustration, we observe the prominent negative correlation between PURE and VIPL (e.g., VIPL v1, VIPL v5), and such disharmony persists even within the VIPL dataset (e.g., v1 and v2). Conceivably, these conflicts can be more prevalent and difficult to identify outside controlled laboratory environments.
% hindering the reasonable definition of domain labels.

\noindent \textbf{Instance-wise gradient harmony}. The observed conflicts in gradients suggest that the model training is hindered by irrelevant attributes and fails to capture the inherent pattern extraction of rPPG. To mitigate this, we propose to harmonize the training process by resolving these conflicts in the parameter hyperplane, which has been shown to be effective in existing works \cite{GradientSurgery, DomainGS}. However, as observed in Fig. \ref{Conflict_domain_pre}, these conflicts are pervasive throughout the dataset, rendering traditional domain labels (e.g., dataset classes) too coarse-grained to represent the clusters of the same attribute \cite{NEST}. Therefore, we select to perform gradient surgery at the instance level, i.e., view each instance as a single cluster. Specifically, before optimizing the model, we first calculate the backward gradient $\mathcal{G}_i$ for each instance $x_i$ ($i \in [0, N-1]$) in the batch of instances $x_0, x_1, \cdots, x_{N-1}$. Then, we neutralize the conflicting portions of the collected backward gradients by obeying the following rule: if two gradients $\mathcal{G}_i$ and $\mathcal{G}_j$ are negatively related (i.e., their cosine similarity is negative), then the gradient of each instance is projected onto the normal hyperplane of the gradient of the other instance, as shown in Equation. (\ref{gradient projection}).

\begin{equation}
    \mathcal{G}_i = \mathcal{G}_i - \frac{\mathcal{G}_i \cdot \mathcal{G}_j}{||\mathcal{G}_j||{2}}\mathcal{G}_j
    \label{gradient projection}
\end{equation}

\noindent where $||\mathcal{G}_j||$ represent the norm of $\mathcal{G}_j$. This operation restricts the conflicting portions of the gradients between two instances while preserving the common parts. And we iteratively apply equation (\ref{gradient projection}) to each gradient pair $\mathcal{G}_i$ and $\mathcal{G}_j$ ($i,j \in [0, N-1]$, $i \neq j$), as outlined in Algorithm \ref{HAPPY-ag} (lines 4-9). By doing so, we thus filter out irrelevant external attributes within the training batch and retain the common information (rPPG), which allows the overall optimization direction to cover as many instances in the batch as possible. Consequently, the deep learning model can learn a generalized way of extracting rPPG signals under different training domains.

\begin{figure}
    \centering
    \includegraphics[width=7.5cm]{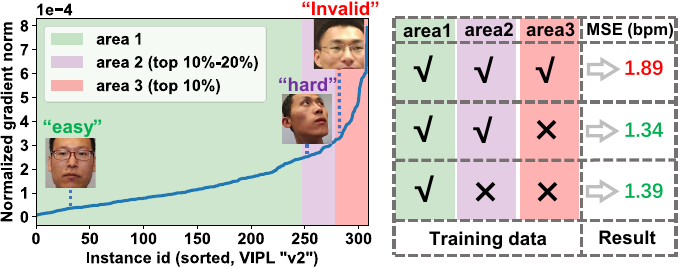}
    \vspace{-4mm}
    \caption{Potential invalid instances (i.e., those with larger gradient norms, area 3) can be disharmonious with others, leading to worse performance of the rPPG model. The involved training instances are collected from task v2 in VIPL \cite{VIPL-HR} database, the testing sets are PURE \cite{PURE} and UBFC \cite{UBFC}.}
    \label{OOD-gradients}
    \vspace{-6mm}
\end{figure}

\noindent \textbf{Global gradient harmony}. The rPPG signal is susceptible to external noise, and in extreme cases, certain instances may be considered invalid due to the overwhelming presence of noise. This phenomenon results in a wide range of difficulty levels among training instances, which is not conducive to the model learning \cite{five_negative, GHM}. The presence of these invalid instances poses a challenge for our harmonious hyperplane optimization, as they may introduce the distorted normal hyperplanes during the gradient projection step (eq. (\ref{gradient projection})), thus compromising the integrity of the harmonized gradient. As illustrated in Fig. \ref{OOD-gradients}, restricting these potential invalid instances (i.e., samples with abnormal gradient norms \cite{GHM}) can significantly upgrade the performance of the rPPG method \cite{CAN}.

In light of this, we aim to mitigate the negative effects of these invalid instances by sifting out the out-of-distribution (OOD) instances during the optimization, as shown in Algorithm 2 (line 3, 9). Specifically, we introduce a global queue $Queue$ with max length $L$ (150 by default) to record the backward gradients of previous instances. Each time before stepping the optimization, we compare the current backward gradients $\mathcal{G}_0,\mathcal{G}_{1},\cdots,\mathcal{G}_{N-1}$ with the $Queue$. If any instance (for example, $\mathcal{G}_1$) in the batch exhibits a significantly larger gradient (larger than 95$\%$ gradients, i.e., ${Top}_{(5\%)}(Queue)$ over the norm in $Queue$), we temporally remove the instance (i.e., set it to zero vector) from the current turn of optimization. Noteworthy, the removed instance will not be permanently discarded. With the rPPG model continuing to learn over time, it can gradually identify and utilize these previous invalid instances to handle high-difficulty tasks. Thus we can globally select the most suitable instances to ensure the progressive training of the rPPG model.

Based on discussions above, we conclude our harmonious hyperplane optimization in Algorithm \ref{HAPPY-ag}, where $\mathbb I(\cdot)$ is the indicator function. Noteworthy, the loss function $\mathcal{L}(\cdot, \cdot)$ in Algorithm \ref{HAPPY-ag} can vary depending on the training strategy. For the harmonious phase strategy, $\mathcal{L}(\cdot, \cdot)$ can be eq. (\ref{MSE_loss}); for traditional wave-based training strategies \cite{PhysNet, CAN, Deepphys}, $\mathcal{L}(\cdot, \cdot)$ can be negative Pearson loss \cite{STVEN}, etc.

\begin{algorithm} 
	\caption{Harmonious Hyperplane Optimization (DOHA-HHO)} 
	\label{HAPPY-ag} 
	\begin{algorithmic}[1]
            \STATE {\bfseries Input:} DL-based model $f_{\theta}(\cdot): \mathcal{X} \to \mathcal{Y}$, batch size $N$, batch of input data $\{x_{0}, x_{1}, \cdots, x_{N-1}\}$, batch of ground truth label $\{y_{0}, y_{1}, \cdots, y_{N-1}\}$, loss function $\mathcal{L}(\cdot, \cdot)$, threshold $T$, gradient record list $Queue$.
            \STATE $\mathcal{G}_i \leftarrow$ $\nabla_{\theta}\mathcal{L}(f_{\theta}(x_i), y_i)$, $\forall i \in [0, N-1]$
            % \\ $\%$ Global gradient harmony (DOHA-GGH) $\%$
            \STATE $\mathcal{G}_i^{GS} \leftarrow$ $\mathbb I_{(||\mathcal{G}_i||_2<Top_{(T\%)}(Queue))}(\mathcal{G}_i)$, $\forall i \in [0, N-1]$
            % \\ $\%$ Instance-wise gradient harmony (DOHA-IGH) $\%$
            \FOR{$i \in [0, N-1]$}
            \FOR{$j \sim [0, N-1] \setminus i$ in random order}
            \IF{$\mathcal{G}_i^{GS} \cdot \mathcal{G}_j < 0$}
            \STATE Conduct $\mathcal{G}_i^{GS} = \mathcal{G}_i^{GS} - \frac{\mathcal{G}_i^{GS} \cdot \mathcal{G}_j}{||\mathcal{G}_j||^{2}}\mathcal{G}_j$
            \ENDIF
            \ENDFOR
            \ENDFOR
            \STATE Update $f_{\theta}( \cdot )$ via $\Delta \theta = \sum_{i}\mathcal{G}_i^{GS}/ N$ 
            \STATE Update $Queue$ with $||\mathcal{G}_0||_2, ||\mathcal{G}_1||_2, \cdots, ||\mathcal{G}_{N-1}||_2$
	\end{algorithmic}
\end{algorithm}

\begin{table*}
\centering
\caption{Heart rate (HR) estimation results over three simulated unseen scenarios under multiple dataset protocol, $^+$ indicates applying this method (on the Baseline \cite{CAN} by default). The unit of HR is bpm (i.e., beat per minute).}
\vspace{-3mm}
\begin{tabular}{lcccccccccccc}
\toprule[1.5pt]
                  &                      & \multicolumn{3}{c}{VIPL} &  & \multicolumn{3}{c}{UBFC} &  & \multicolumn{3}{c}{PURE} \\ \cline{3-5} \cline{7-9} \cline{11-13} 
\multirow{-2}{*}{Method} &
   &
  MAE$\downarrow$ &
  RMSE$\downarrow$ &
  r$\uparrow$ &
   &
  MAE$\downarrow$ &
  RMSE$\downarrow$ &
  r$\uparrow$ &
   &
  MAE$\downarrow$ &
  RMSE$\downarrow$ &
  r$\uparrow$ \\ \midrule[1pt]
GREEN \cite{GREEN}            &                      & 16.37   & 21.90  & 0.27  &  & 7.68   & 11.15   & 0.55  &  & 3.12   & 7.36   & 0.81   \\
CHROM \cite{CHROM}            &                      & 11.28   & 15.12  & 0.54  &  & 4.88   & 8.71    & 0.83  &  & 2.57   & 6.85   & 0.84   \\
POS \cite{POS}              &                      & 12.94   & 18.73  & 0.38  &  & 4.33   & 8.44    & 0.85  &  & 2.77   & 4.51   & 0.91   \\ \hline
PhysNet \cite{PhysNet} &
   &
  10.43 &
  16.03 &
  0.44 &
   &
  2.03 &
  3.12 &
  0.94 &
  &
  2.54 &
  4.33 &
  0.92 \\
Deepphys \cite{Deepphys} &
  \multicolumn{1}{l}{} &
  12.35 &
  15.48 &
  0.43 &
  \multicolumn{1}{l}{} &
  1.78 &
  2.96 &
  0.95 &
   &
  4.01 &
  6.69 &
  0.85 \\
TS-CAN \cite{CAN}           &                      & 12.12   & 14.12  & 0.48  &  & 1.58   & 2.71    & 0.96  &  & 3.75   & 6.13   & 0.88   \\ \hline
SAM$^+$ \cite{SAM}            & \multicolumn{1}{l}{} & 12.01   & 13.78  & 0.47  &  & 1.67   & 3.35    & 0.95  &  & 3.59   & 6.01   & 0.88   \\
Domain GS$^+$  \cite{DomainGS}             & \multicolumn{1}{l}{} & 11.47   & 14.11  & 0.48  &  & 1.59   & 2.70    & 0.96  &  & 3.61   & 6.21   & 0.87 \\
NEST\cite{NEST}          & \multicolumn{1}{l}{} & 11.01   & 14.33  & 0.49  &  & 1.54   & 3.01    & 0.97  &  & 2.93   & 4.13   & 0.90
\\ \hline
Baseline \cite{CAN}         &                      & 12.12   & 14.12  & 0.48  &  & 1.58   & 2.71    & 0.96  &  & 3.75   & 6.13   & 0.88   \\
DOHA$^+$ (w/o HPS) &                      & 10.73   & 13.71  & 0.53  &  & 1.53   & 2.81    & 0.97  &  & 1.78   & 2.96   & 0.91   \\
DOHA$^+$ (w/o GGH)    &                      & 7.46    & 12.09  & 0.61  &  & 1.76   & 2.95    & 0.96  &  & 1.25   & 2.31   & 0.95   \\
DOHA$^+$ (w/o IGH)    &                      & 7.38    & 11.77  & 0.62  &  & 1.46   & 1.83    & \textbf{0.98}  &  & 1.03   & 1.79   & 0.98   \\
DOHA$^+$ &
   &
  \textbf{6.98} &
  \textbf{10.95} &
  \textbf{0.67} &
   &
  \textbf{1.41} &
  \textbf{1.56} &
  \textbf{0.98} &
   &
  \textbf{0.95} &
  \textbf{1.58} &
  \textbf{0.99} \\ \bottomrule[1.5pt]
\end{tabular}
\label{MainTable}
\vspace{-2mm}
\end{table*}

\vspace{-6mm}
\section{Experiments}
\vspace{-1mm}
\subsection{Datasets}
 % Be careful that videos of the subject 25-27 have the outlier frame rate (i.e. fps), which is about 23 (e.g. others are around 30 fps). 
\textbf{VIPL-HR} \cite{VIPL-HR} is a large-scale dataset for remote physiological measurement, it contains different tasks covering different ambient light conditions and motion intensities on 107 subjects recorded by 4 different devices. \textbf{UBFC} \cite{UBFC} contains 42 videos recorded from 42 subjects respectively. \textbf{PURE} \cite{PURE} contains facial videos recorded from 10 subjects, each subject participates in 6 different tasks (sitting still, talking, four variations of head movement). Like \cite{NEST}, when collecting training instances, we align the frame rate to 30 fps by interpolation.

% \textbf{BUAA} \cite{BUAA} contains videos recorded under 10 different illumination conditions. Like \cite{NEST}, we use videos with illumination greater than 10 lux in the experiment.

\vspace{-2.5mm}
\subsection{Implementation Details}
\textbf{Training Details}. All experiments can be conducted on a single GeForce RTX 3090 graphic card. When processing facial videos, we utilize the ``mmod face detector" \cite{mmod} from the ``dlib" python package to detect human faces. We then resize each detected region to a shape of $[151 \times 151]$. For data augmentation, we apply RandomCrop by randomly selecting a $[131 \times 131]$ \footnote{We sightly adjust the entrance of involved network pipeline \cite{Deepphys, CAN, PhysNet} to fit the input size, the detailed adjustment refers to Appendix \ref{Appendix Network adjustment}.} region from the original $[151 \times 151]$ frame for each video. We eliminated the temporal padding in the baseline \cite{CAN} to ensure the quality of the information on collected temporal windows. The training video clip length is set to 75, the default batch size is 4, and the total training epoch is 20. We employ the ``adam" optimizer \cite{adam} with a cosine learning rate (lr) scheduler \cite{SGDR} (initial lr 5e-4, min lr 1e-6) to train DL models. During testing, the input video clip length is set to 300, which follows the same protocol \cite{RhythmNet, PhysFormer}. The default sifting threshold $T$ in DOHA-GGH is set as 5, and the length of the $Queue$ is 150. Before generating the self-similarity physiological map shown in Fig. \ref{SSP}, we first filter the ground truth wave (using a band-pass filter with $[0.7, 3.5]$ Hz). We use the spatially averaged intermediate output $x_{[T \times 1 \times 1]}$ of the last hidden layer in the DL model \cite{CAN, Deepphys, PhysNet} to generate the output self-similarity physiological map $\hat{\mathcal{R}}$. The default sliding window length $L_w$ is set to 17. We use equation (\ref{MSE_loss}) as the loss function $\mathcal{L}(\cdot, \cdot)$ to optimize the DL model with DOHA-HPS and follow the original traditional loss strategies of \cite{CAN, PhysNet, Deepphys, RhythmNet} when training without DOHA-HPS.

\noindent \textbf{Performance Metrics}. We focus on the performance of heart rate (HR),  heart rate variability (HRV), and respiration frequency (RF) estimations. Following previous works \cite{RhythmNet,PhysFormer,Dual-GAN, NEST}, we adopt mean absolute error (MAE), root mean square error (RMSE), and Pearson correlation coefficient (r) for evaluation.

\vspace{-4mm}
\subsection{Improvement on Multiple Dataset Training}
\subsubsection{Improvement on HR Estimation}
We evaluated the HR estimation performance on unseen scenarios with three datasets: VIPL \cite{VIPL-HR}, PURE \cite{PURE}, and UBFC \cite{UBFC}. To simulate the unseen scenarios, we train the model on two datasets and then test it on the remaining dataset. We first compare DOHA against three traditional rPPG methods (GREEN \cite{GREEN}, CHROM \cite{CHROM}, and POS \cite{POS}) using the pyVHR package \cite{pyVHR}. Then we re-implement three classical rPPG DL methods, i.e., DeepPhys \cite{Deepphys}, PhysNet \cite{PhysNet}, and TS-CAN \cite{CAN}. As reported in \cite{NEST}, TS-CAN does not perform well against domain shift, so we use it as the baseline to evaluate the improvement of DOHA. Finally, we compare our method against three domain generalization (DG) methods, including SAM \cite{SAM}, DomainGS \cite{DomainGS} and NEST \cite{NEST}.

\begin{table}
\centering
\caption{Universality verification of DOHA on other widely-used rPPG methods. $^{+}$ means adopting DOHA on this method.}
\vspace{-3mm}
\begin{tabular}{lccc}
\toprule[1.5pt]
\multirow{2}{*}{Method}                  & \multicolumn{3}{c}{VIPL}                           \\ \cline{2-4} 
                                         & MAE$\downarrow$ & RMSE$\downarrow$ & r$\uparrow$   \\ \midrule[1pt]

RhythmNet\cite{RhythmNet} & 8.97            & 12.16            & 0.49          \\
RhythmNet$^+$                               & \textbf{7.75}   & \textbf{10.97}   & \textbf{0.63} \\ \hline
Deepphys\cite{Deepphys} & 12.35           & 15.48            & 0.43          \\
Deepphys$^+$                                & \textbf{7.48}   & \textbf{11.04}   & \textbf{0.61} \\ \hline
PhysNet\cite{PhysNet}   & 10.43           & 16.03            & 0.44          \\
PhysNet$^+$                                 & \textbf{7.33}   & \textbf{12.25}   & \textbf{0.62} \\ \bottomrule[1.5pt]
\end{tabular}
\label{MSDG}
\vspace{-8mm}
\end{table}

\noindent \textbf{The performance of traditional methods.} Tab. \ref{MainTable} shows that traditional methods can retain acceptable performance under simple scenarios like UBFC and PURE, but their performance significantly deteriorates in more challenging scenarios such as VIPL \cite{VIPL-HR}. This suggests that these traditional methods that rely on static prior knowledge \cite{POS} may not be competent for handling complex tasks and thus are vulnerable to external factors such as head motion and illumination variation.

\begin{table*}
\centering
\caption{The improvement of DOHA over HRV and RF estimation, $^+$ indicates adopting this method to the Baseline \cite{CAN}.}
\vspace{-3.5mm}
\resizebox{2\columnwidth}{!}{
\begin{tabular}{cccccccccccccccccc}
\toprule[2pt]
\multirow{2}{*}{Target} &
  \multirow{2}{*}{Method} &
   &
  \multicolumn{3}{c}{LF-(u.n)} &
   &
  \multicolumn{3}{c}{HF-(u.n)} &
   &
  \multicolumn{3}{c}{LF/HF} &
   &
  \multicolumn{3}{c}{RF(Hz)} \\ \cline{4-6} \cline{8-10} \cline{12-14} \cline{16-18} 
 &
   &
   &
  std$\downarrow$ &
  RMSE$\downarrow$ &
  r$\uparrow$ &
   &
  std$\downarrow$ &
  RMSE$\downarrow$ &
  r$\uparrow$ &
   &
  std$\downarrow$ &
  RMSE$\downarrow$ &
  r$\uparrow$ &
   &
  std$\downarrow$ &
  RMSE$\downarrow$ &
  r$\uparrow$ \\ \midrule[1pt]
\multirow{5}{*}{UBFC} &
  GREEN \cite{GREEN} &
   &
  0.246 &
  0.305 &
  0.1849 &
   &
  0.246 &
  0.305 &
  0.1849 &
   &
  1.094 &
  1.275 &
  0.1085 &
   &
  0.090 &
  0.091 &
  0.005 \\
 &
  CHROM \cite{CHROM} &
   &
  0.226 &
  0.292 &
  0.317 &
   &
  0.226 &
  0.292 &
  0.317 &
   &
  1.020 &
  1.120 &
  0.275 &
   &
  0.084 &
  0.085 &
  0.079 \\
 &
  POS \cite{POS} &
   &
  0.234 &
  0.299 &
  0.250 &
   &
  0.234 &
  0.299 &
  0.250 &
   &
  1.010 &
  1.228 &
  0.253 &
   &
  0.084 &
  0.086 &
  0.046 \\
 &
  Baseline \cite{CAN} &
   &
  0.216 &
  0.259 &
  0.396 &
   &
  0.216 &
  0.259 &
  0.396 &
   &
  1.014 &
  1.186 &
  0.403 &
   &
  0.087 &
  0.087 &
  0.047 \\
 &
  DOHA$^+$ &
   &
  \textbf{0.212} &
  \textbf{0.258} &
  \textbf{0.429} &
  \textbf{} &
  \textbf{0.212} &
  \textbf{0.258} &
  \textbf{0.429} &
  \textbf{} &
  \textbf{1.007} &
  \textbf{1.178} &
  \textbf{0.418} &
  \textbf{} &
  \textbf{0.083} &
  \textbf{0.083} &
  \textbf{0.101} \\ \hline
\multirow{5}{*}{PURE} &
  GREEN \cite{GREEN}&
   &
  0.279 &
  0.314 &
  0.061 &
   &
  0.279 &
  0.314 &
  0.061 &
   &
  1.128 &
  1.286 &
  0.0848 &
   &
  0.084 &
  0.085 &
  -0.001 \\
 &
  CHROM \cite{CHROM}&
   &
  0.245 &
  0.305 &
  0.172 &
   &
  0.245 &
  0.305 &
  0.172 &
   &
  1.086 &
  1.287 &
  0.115 &
   &
  0.091 &
  0.093 &
  0.004 \\
 &
  POS \cite{POS}&
   &
  0.272 &
  0.298 &
  0.008 &
   &
  0.272 &
  0.298 &
  0.008 &
   &
  1.002 &
  1.107 &
  0.052 &
   &
  0.086 &
  0.087 &
  -0.021 \\
 &
  Baseline \cite{CAN}&
   &
  0.237 &
  0.266 &
  0.288 &
   &
  0.237 &
  0.266 &
  0.288 &
   &
  1.009 &
  1.110 &
  0.279 &
   &
  0.084 &
  0.086 &
  0.003 \\
 &
  DOHA$^+$ &
   &
  \textbf{0.236} &
  \textbf{0.259} &
  \textbf{0.307} &
  \textbf{} &
  \textbf{0.236} &
  \textbf{0.259} &
  \textbf{0.307} &
  \textbf{} &
  \textbf{0.951} &
  \textbf{1.041} &
  \textbf{0.354} &
  \textbf{} &
  \textbf{0.082} &
  \textbf{0.083} &
  \textbf{0.099} \\ \bottomrule[1.5pt] 
\end{tabular}
}
\label{HRV_table}
\vspace{-3mm}
\end{table*}

\noindent \textbf{The performance of DL-based rPPG methods.} As shown in Tab. \ref{MainTable}, the involved DL methods, i.e., PhysNet \cite{PhysNet}, DeepPhys \cite{Deepphys} and TS-CAN \cite{CAN}, exhibit merely comparable performance to traditional methods like POS \cite{POS} when facing all three unseen scenarios. Such result suggests that the baseline \cite{CAN} has difficulty in learning generalized knowledge to analyze the rPPG signal under the traditional training setting (i.e. following the loss function \cite{CAN} of the baseline), and this drawback is amplified under the multiple dataset training scenario where the domain conflicts within the training database are more severe.

\begin{table}
\centering
\caption{HR estimation results over two unseen domains (VIPL and PURE), the training dataset is UBFC \cite{UBFC}.}
\vspace{-3.5mm}
\resizebox{1\columnwidth}{!}{
\begin{tabular}{lcccccccc}
\toprule[1.5pt]
\multicolumn{1}{c}{\multirow{2}{*}{Method}} &  & \multicolumn{3}{c}{VIPL}                         &  & \multicolumn{3}{c}{PURE}                         \\ \cline{3-5} \cline{7-9} 
\multicolumn{1}{c}{}                        &  & MAE$\downarrow$ & RMSE$\downarrow$ & r$\uparrow$ &  & MAE$\downarrow$ & RMSE$\downarrow$ & r$\uparrow$ \\ \midrule[1pt]
GREEN\cite{GREEN}      &  & 16.37 & 21.90 & 0.27 &  & 3.12 & 7.36 & 0.81 \\
CHROM\cite{CHROM}      &  & 11.28 & 15.12 & 0.54 &  & 2.57 & 6.85 & 0.84 \\
POS\cite{POS}          &  & 12.94 & 18.73 & 0.38 &  & 2.77 & 4.51 & 0.91 \\ \hline
Baseline\cite{CAN} &  & 12.21 & 19.89 & 0.44 &  & 2.67 & 7.63 & 0.80 \\
DOHA$^+$              &  & \textbf{7.68}  & \textbf{12.69} & \textbf{0.60} &  & \textbf{1.03} & \textbf{1.39} & \textbf{0.97} \\ \bottomrule[1.5pt]
\end{tabular}}
\label{SSDG}
\vspace{-5mm}
\end{table}

\noindent \textbf{The performance of DG methods}. According to Tab. \ref{MainTable}, none of the three domain generalization (DG) methods (i.e. SAM \cite{SAM}, Domain GS \cite{DomainGS}, and NEST \cite{NEST}) show significant performance improvement over the baseline method \cite{CAN}. We argue that this is due to their limited ability to address the underlying domain conflicts present in the training datasets. SAM's approach \cite{SAM} of achieving a smooth loss surface may lead to a sub-optimal solution that only covers certain parts of the training scenarios. The domain conflicts in rPPG databases are more complicated, which may exceed the capabilities of SAM to address effectively. Domain-based Gradient Surgery \cite{DomainGS}, which relies on coarse-grained domain labels, may not accurately capture the complex domain differences in rPPG databases, making it challenging to neutralize the domain conflicts thoroughly. Although NEST \cite{NEST} performs better than the other two DG methods, it still does not fully address the label conflicts, which limits its further improvement.

\noindent \textbf{The performance of DOHA-HPS}. As demonstrated in Tab. \ref{MainTable}, adopting our harmonious phase strategy can lead to a substantial improvement in performance for the baseline method \cite{CAN}. This suggests that the harmonious phase strategy can provide a more appropriate label representation that assists the baseline to learn domain-invariant knowledge \cite{AD, krueger2021out, parascandolo2020learning, NCDG}. By mitigating the phase delay differentials and focusing on the intrinsic relationship within the physiological signal, the new label representation thus serves as a better guide for the model, enabling it to achieve more generalized performance even in unseen scenarios.

\begin{table}
\centering
\caption{DOHA can enhance the performance of baseline \cite{CAN} to the level of existing state-of-the-art rPPG methods (i.e., PhysFormer\cite{PhysFormer}, RhythmNet\cite{RhythmNet}) on VIPL dataset.}
\vspace{-3.5mm}
\resizebox{0.65\columnwidth}{!}{
\begin{tabular}{lccc}
\toprule[1.5pt]
\multirow{2}{*}{Method}                  & \multicolumn{3}{c}{VIPL}                           \\ \cline{2-4} 
                                         & MAE$\downarrow$ & RMSE$\downarrow$ & r$\uparrow$   \\ \midrule[1pt]
% GREEN \cite{GREEN} & 16.37 & 21.90 & 0.27 \\
CHROM \cite{CHROM} & 11.28 & 15.12 & 0.54 \\
POS \cite{POS} & 12.94 & 18.73 & 0.38 \\
\hline
RhythmNet \cite{RhythmNet}           & 5.30            & 8.14             & 0.76          \\
TS-CAN \cite{CAN}            & 8.08            & 14.83            & 0.63          \\
Physformer \cite{PhysFormer}          & 4.97            & 7.79             & 0.78          \\
Dual-GAN \cite{Dual-GAN}            & 4.93            & 7.68             & 0.81          \\ 
% NEST \cite{NEST}            & \textbf{4.76}            & \textbf{7.51}             & \textbf{0.84}         \\ 
\hline
Baseline \cite{CAN}            & 8.08            & 14.83            & 0.63          \\
DOHA+ (w/o HPS) & 7.07            & 11.17            & 0.65          \\
DOHA+ (w HPS)   & 5.01            & 8.49             & 0.78          \\
DOHA+               & \textbf{4.87}   & \textbf{7.64}    & \textbf{0.83} \\ \bottomrule[1.5pt]
\end{tabular}
}
\label{Inner}
\vspace{-4mm}
\end{table}

\noindent \textbf{The performance of DOHA-HHO}. In Tab. \ref{MainTable}, we observe that the harmonious hyperplane optimization improves the performance of the baseline method \cite{CAN}, both with and without the harmonious phase strategy. This result indicates that our DOHA-HHO can effectively address the attribute conflicts among rPPG datasets. As analyzed, the irrelevant attribute conflicts among instances exist in a fine-grained way, where the instance-wise gradient surgery can effectively cover and mitigate them. Moreover, the global gradient harmony helps maintain a balanced overall training data difficulty, thus preventing the model from being excessively influenced by instances that may lead to learning irrelevant attributes.

\vspace{-2.5mm}
\subsubsection{Improvement on HRV and RF}
Following \cite{NEST, PhysFormer, McDuff_HRV}, we conduct experiments to examine the effect of DOHA-HHO on HRV (LF, HF, and LF/HF) and RF estimation using the PURE \cite{PURE} and UBFC \cite{PURE} datasets. Likely, we train the model on one dataset and then test it on another to simulate the unseen scenario. As shown in Tab. \ref{HRV_table}, DOHA provides a more dominant performance for the baseline method \cite{CAN}, which indicates that our methodology not only extracts periodic features but also maximally preserves the temporal physiological information.
 
\vspace{-2.5mm}
\subsubsection{Universality Verification}

DOHA is plug-and-play, making it able to take effect for both end-to-end methods \cite{Deepphys,PhysNet} and hand-manual (i.e. non end-to-end) methods \cite{RhythmNet}. To prove this, we train these models \cite{Deepphys, PhysNet, RhythmNet} on UBFC, PURE, and BUAA \cite{BUAA} and test the performance of DOHA on VIPL. As shown in Tab. \ref{MSDG}, DOHA consistently improves the performance of all DL-based models. This is because, in DOHA, the harmonious hyperplane optimization introduces constraints to the gradient, which is a fundamental component of the DL training process. Additionally, the harmonious representation strategy can provide a more stable and universal physiological representation for these methods. These properties allow DOHA to take effect in various rPPG training scenarios.

\vspace{-3mm}
\subsection{Improvement on Single Dataset Training}
\textbf{Single-Source Domain Generalization}. Under the worst case, only a single database is available for rPPG model training, which is termed as Single-Source domain generalization \cite{NEST}. To verify the effect of DOHA under this situation, we conduct the experiment by training the baseline \cite{CAN} on UBFC and then test its performance on PURE and VIPL. The results are shown in Tab. \ref{SSDG}.

\noindent \textbf{Intra-Dataset Testing}. Following the experimental setup of \cite{PhysFormer, RhythmNet, Dual-GAN}, we conduct experiments to verify the effect of DOHA under the intra-dataset scenario using a 5-fold subject exclusive cross-validation protocol on the VIPL dataset \cite{VIPL-HR}. The baseline is TS-CAN \cite{CAN}. The performance of DOHA and other state-of-the-art rPPG methods \cite{PhysFormer, Dual-GAN} are presented in Tab. \ref{Inner}.

\noindent \textbf{Analysis}. According to the results, DOHA improves the performance of the baseline \cite{CAN} under both two single dataset training scenarios. This confirms our assumption that the domain conflict phenomenon can also be pervasive within the single rPPG database \cite{VIPL-HR}. In addition, the performance improvement by harmonious phase strategy (in Tab. \ref{Inner}) emphasizes that the information encoded in the self-similarity physiological map can effectively enhance the model's learning of inherent facial cardiac feature analysis.

\vspace{-2.5mm}
\subsection{Further Discussion}
\begin{figure}
    \centering
    \subfloat[$L_{win}$]{\includegraphics[height=2.1cm]{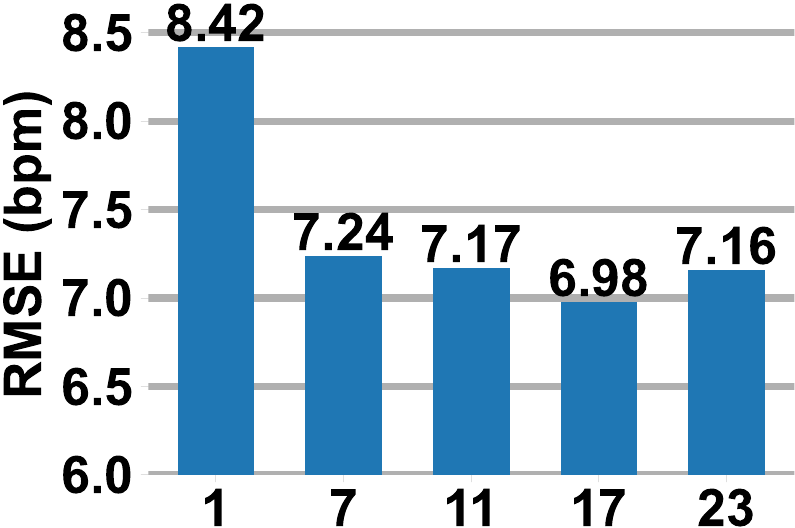}} \hfill
    \subfloat[$T$ ($\%$)]{\includegraphics[height=2.1cm]{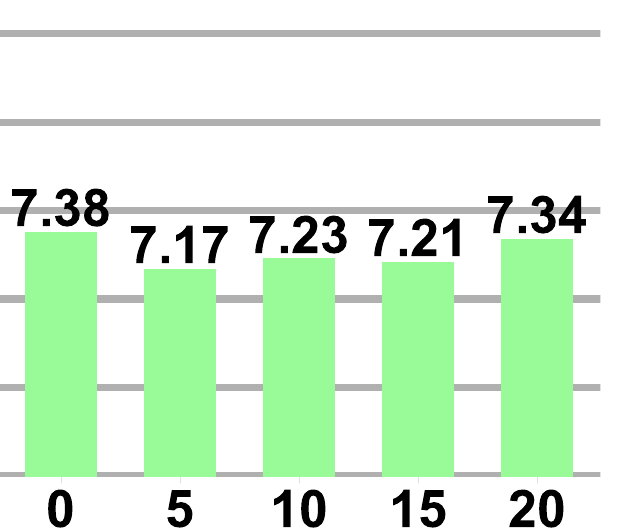}} \hfill
    \subfloat[Batch size]{\includegraphics[height=2.1cm]{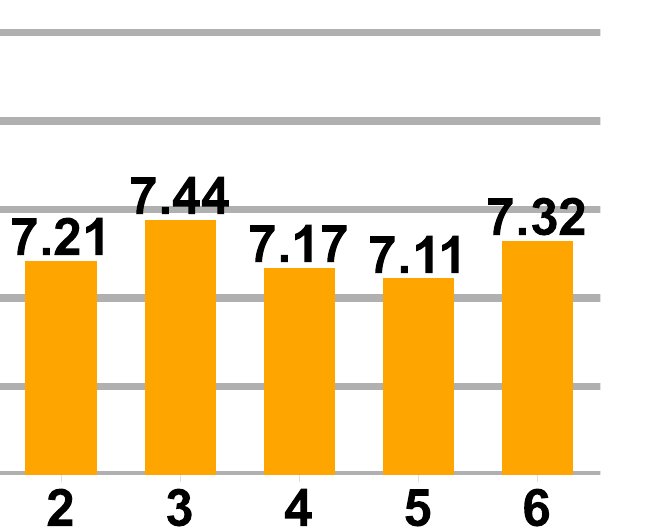}}
    \vspace{-4mm}
    \caption{Ablation study on hyperparameters of DOHA.}
    \label{Ablation_WB}
    \vspace{-4mm}
\end{figure}

\textbf{The impact of $L_{win}$}. In DOHA-HPS, the sliding window length $L_{win}$ determines the temporal perception to the physiological signal. We conduct an ablation study to seek the optimal $L_{win}$. As depicted in Fig. \ref{Ablation_WB} (a), we observed that generally a longer sight (i.e., larger $L_{win}$) results in improved performance. This phenomenon can be attributed to the fact that a longer sight allows the model to analyze the temporal information with more temporal context, thus enhancing its ability to resist burst noise (e.g., sudden, fierce subject's movement). However, a longer $L_{win}$ also introduces more temporal consumption, which can have a negative impact and needs to be carefully considered.

\noindent \textbf{The impact of $T$}. In DOHA-GGH, threshold $T$ balances the global difficulty of participating instances. As shown in Fig. \ref{OOD-gradients} and \ref{Ablation_WB} (b), sightly restricting those out-of-distribution instances can properly enhance the training effect, which matches the experimental results in other fields \cite{GHM, five_negative}. Noteworthy, for databases with higher diversity in difficulty levels (e.g., VIPL v2), the effect of DOHA-GGH can be more pronounced. But larger $T$ reduces the utilization of the dataset, which may hurt the generalization of model \cite{Data-aug}.

\begin{table}
\centering
\caption{The impact of the projection layer in DOHA-HPS, where the ``Identity'' equals removing the projection layer.}
\vspace{-3mm}
\resizebox{0.7\columnwidth}{!}{
\begin{tabular}{c|c|cccc}
\toprule[1.5pt]
\multirow{2}{*}{\begin{tabular}[c]{@{}c@{}}Projection\\ Layer\end{tabular}} & \multirow{2}{*}{Structure} &  & \multicolumn{3}{c}{VIPL} \\ \cline{4-6}
 &
   &
   &
  MAE$\downarrow$ &
  RMSE$\downarrow$ &
  r$\uparrow$
   \\ \hline
Identity &
  - &
   & 7.82
   & 13.19
   & 0.57
   \\ \hline
\multirow{2}{*}{\begin{tabular}[c]{@{}c@{}}Linear\\ Layer\end{tabular}} &
  {[}88$\times$88{]} &
   & 6.98
   & \textbf{10.95}
   & 0.67
   \\ \cline{2-2}
 &
  {[}88$\times$352{]} &
   & \textbf{6.94}
   & 11.65
   & \textbf{0.69}
   \\ \hline
BottleNeck &
  \begin{tabular}[c]{@{}c@{}}{[}88$\times$88,\\ 88$\times$352,\\ 88$\times$88{]}\end{tabular} &
   & 7.56
   & 12.67
   & 0.60
   \\ \bottomrule[1.5pt]
\end{tabular}}
\label{Projection Ablation}
\vspace{-4.5mm}
\end{table}

\noindent \textbf{Discussion on the projection layer}. In the self-similarity attention module, a projection layer is added to interpret the feature of temporal windows. We verify the effect of different projection layers (i.e., identity, single linear layer, and bottleneck). In the ablation study, the protocol agrees with Tab. \ref{MainTable}, and the testing set is VIPL \cite{VIPL-HR}. Results in Tab. \ref{Projection Ablation} indicate that a proper projection layer can significantly improve the performance. This is because the projection layer can extract inherent physiological information among different instances while alleviating their differentials. For example, some instances in UBFC \cite{UBFC} is experimentally proven to convey the diastolic peak, while other instances may not have this property. Under this circumstance, a proper projection layer can effectively align these instances, leading to better generalization. 

\noindent \textbf{Discussion on computing resource}. DOHA's time consumption (training stage) is acceptable due to the lightweight model structure \cite{PhysNet, CAN} and small batch size (e.g., 4 in \cite{PhysFormer}) of rPPG network training. Besides, the ablation study in Fig. \ref{Ablation_WB} (c) demonstrates that DOHA can work reliably with different widely-used batch sizes. Furthermore, DOHA's time consumption during inference is either negligible. We describe these time consumptions in detail in the Appendix. \ref{Time consumption of DOHA}.

% \noindent \textbf{Discussion on computing resource}. In common case \cite{DomainGS}, the gradient surgery works at the cost of GPU memory and time consumption. Fortunately, this negative effect can be mitigated in the majority of rPPG tasks \cite{PhysNet, CAN, Deepphys, PhysFormer}. This can be attributed to the lightweight model structure \cite{Deepphys, CAN, PhysNet} and small batch size (e.g., 4 in \cite{PhysFormer}) of rPPG network training. For example, many rPPG models \cite{PhysNet, Deepphys, CAN} take up merely 2 megabytes or less storage, making the gradient operation feasible. Additionally, the ablation study presented in Fig. \ref{Ablation_WB} (c) demonstrates that DOHA can work reliably with different widely-used batch sizes. This further supports the convenience and generalization of the DOHA framework.

\begin{figure}
    \centering
    \includegraphics[height=3.05cm]{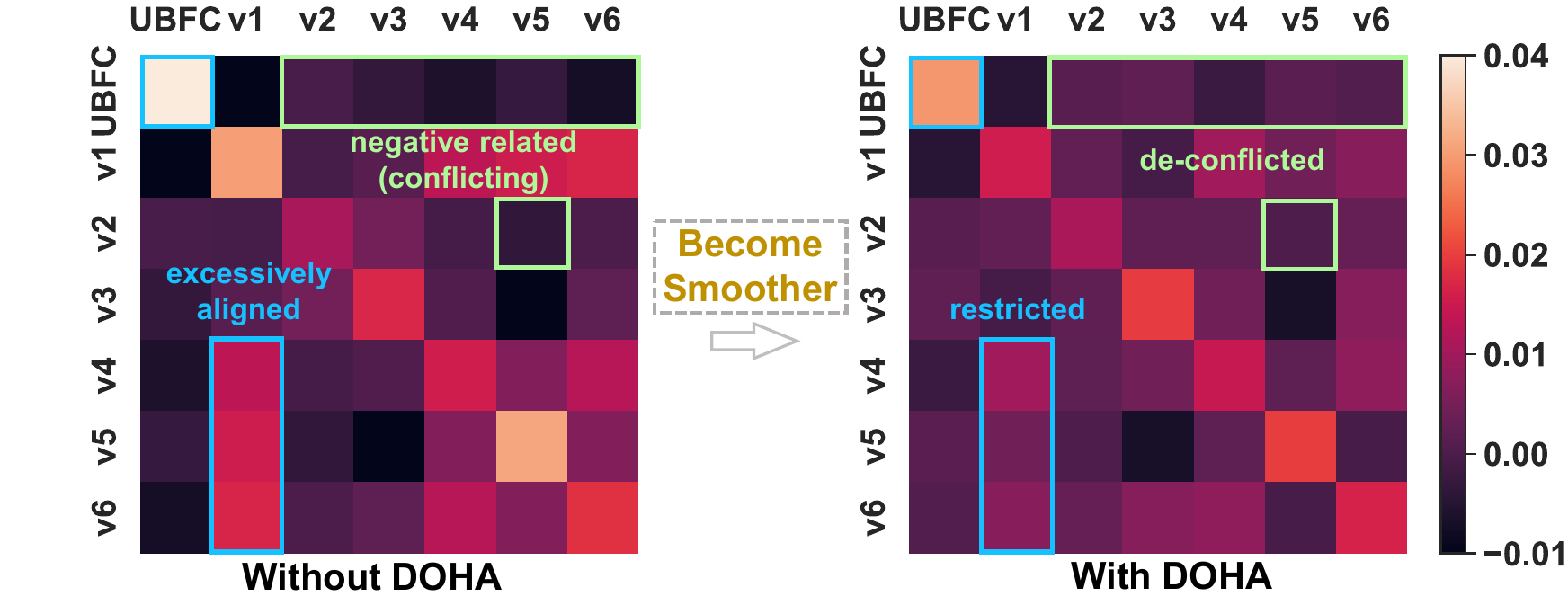}
    \vspace{-3.5mm}
    \caption{Visualization on the effect of DOHA: the relationship among the instances from different attributes becomes smoother \cite{SAM, SAR} (i.e. more uniformed).}
    \label{Visualization}
    \vspace{-5mm}
\end{figure}

\noindent \textbf{Visualization}. To illustrate the effect of DOHA, we visualize the relationship among the gradients of instances in VIPL and UBFC, with the baseline \cite{CAN} trained on these databases. Results in Fig. \ref{Visualization} demonstrate that DOHA helps the baseline model find a smoother solution in the parameter hyperplane: (1) The gradients corresponding to similar attributes (highlighted by the blue box) exhibit restricted extreme similarity, indicating that the excessive training towards these attributes, which can lead to overfitting, is mitigated. (2) Conflicts among different attributes (highlighted by the green box) are alleviated, as evidenced by the reduced negative correlation. Such a phenomenon suggests that DOHA promotes a more balanced representation of different attributes.

% \vspace{-7mm}

\vspace{-2mm}
\section{Conclusion}
In this paper, we define and analyze the domain conflict problem in the rPPG task. We then introduce the DOmain-HArmonious framework (DOHA) to alleviate this problem. DOHA consists of two components, one that neutralizes the unknown phase differentials among instances, and another that alleviates the irrelevant attribute conflicts among different training scenarios. By addressing these conflicts, DOHA enables the rPPG model to focus more on the underlying physiological nature present in the instances, leading to improved generalization performance on unseen scenarios. Importantly, DOHA can be easily integrated into existing DL rPPG methods without requiring additional knowledge such as the domain label. We have conducted extensive experiments to evaluate the effect of DOHA.

\vspace{-2mm}
\begin{acks}
This work was supported in part by Jiangsu Provincial Social Development Key R$\&$D Program (BE2020685).
\end{acks}

\bibliographystyle{ieee_fullname}
\bibliography{egbib}

\appendix

\section{Details on involved databases}
\label{Details on involved databases}
This section demonstrates the details of different external attributes of the involved databases.

(1) \textbf{VIPL-HR}.
Subjects in VIPL-HR database mainly have yellow skin. Each subject participates in 9 different tasks (namely v1-v9). In our main text, we utilize 6 tasks (i.e. v1-v6) for illustration. The main attributes of these 6 tasks are described as follows:

\textbf{v1}. The head movement of the subject is small, and the illumination condition is normal (i.e. as usual).

\textbf{v2}. The head movement of the subject is large (e.g., fierce head shaking), and the illumination condition is normal.

\textbf{v3}. The head movement of the subject is middle (e.g., speaking), and the illumination condition is normal.

\textbf{v4}. The head movement of the subject is small, and the illumination condition is high (i.e. bright).

\textbf{v5}. The head movement of the subject is small, and the illumination condition is low (i.e. dark).

\textbf{v6}. The head movement of the subject is small, and the illumination condition is normal (i.e. as usual). And the subject sits relatively far (about 2m) in front of other tasks (about 0.5m).

\textbf{v7}. The head movement of the subject is small, and the illumination condition is normal (i.e. as usual). And the subject's heart rate is high as he/she is after outdoor sports.

\textbf{v8}. The head movement of the subject is small, and the illumination condition is normal (i.e. as usual). And the subject holds the camera to record facial videos, which may introduce motion because of the head (which holds the camera) movement.

\textbf{v9}. The head movement of the subject is large, and the illumination condition is normal (i.e. as usual).  And the subject holds the camera to record facial videos, which may introduce motion because of the head (which holds the camera) movement.

(2) \textbf{UBFC}.
Subjects in UBFC database mainly have white skin. The video numbers in UBFC is far smaller than those in VIPL, adding the average difficulty of the tasks in UBFC is relative low and balanced, we view them as a whole group in the illustration (Fig. 8) in the main text. 

(3) \textbf{PURE}.
Subjects in PURE database mainly have white skin, and the illumination is comparable low. The video numbers in PURE is far smaller than those in VIPL, adding the average difficulty of the tasks in UBFC is relative low and balanced, we view them as a whole group in the illustration (Fig. 5) in the main text. 

\section{Network adjustment}
\label{Appendix Network adjustment}
\begin{figure}[htbp]
    \vspace{-2mm}
    \centering
    \includegraphics[width=8cm]{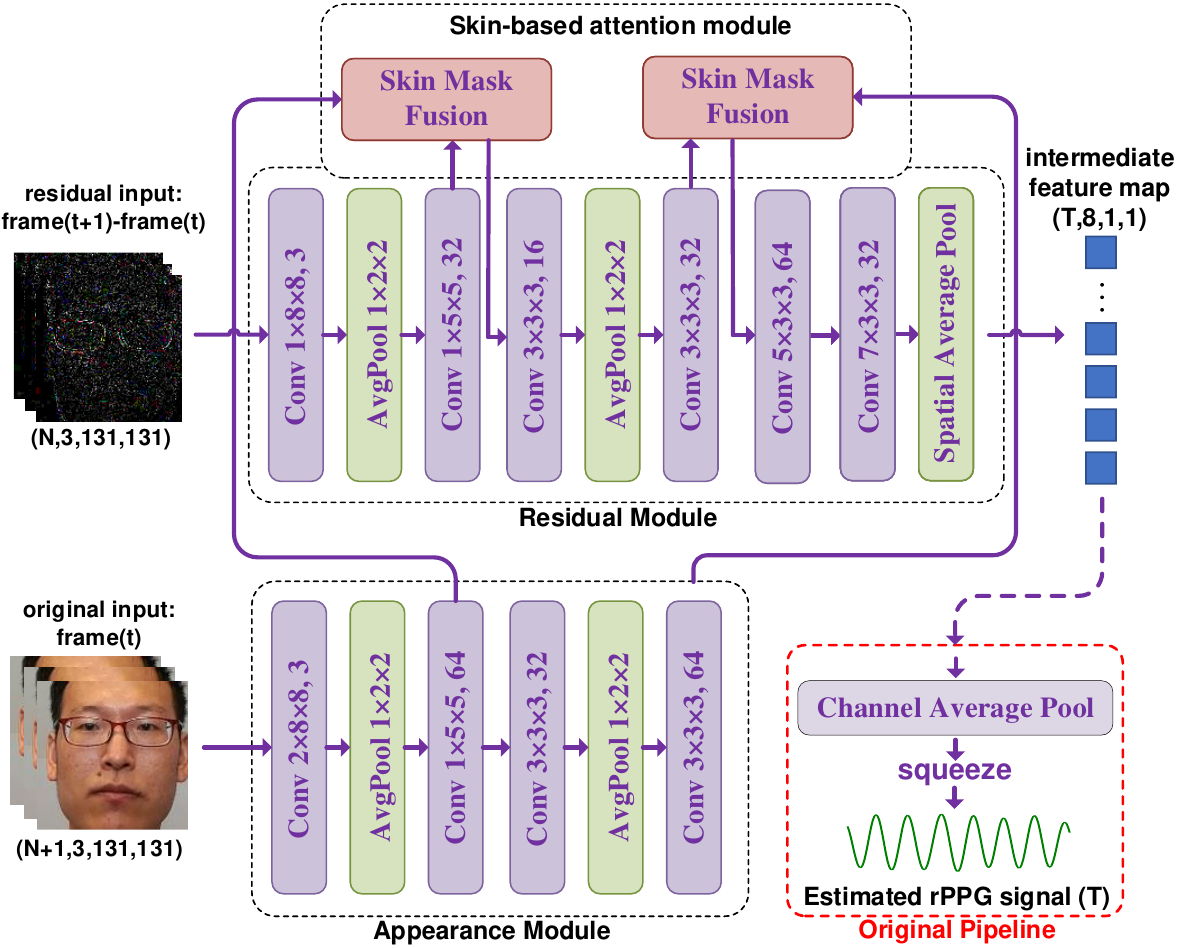}
    \caption{The adjusted baseline (i.e. TS-CAN) in our experiment. The ``Skin Mask Fusion'' can refer to original paper, which is the same.}
    \label{Appendix: Original Baseline}
    \vspace{-5mm}
\end{figure}

\begin{figure*}[htbp]
    \centering
    \includegraphics[width=15cm]{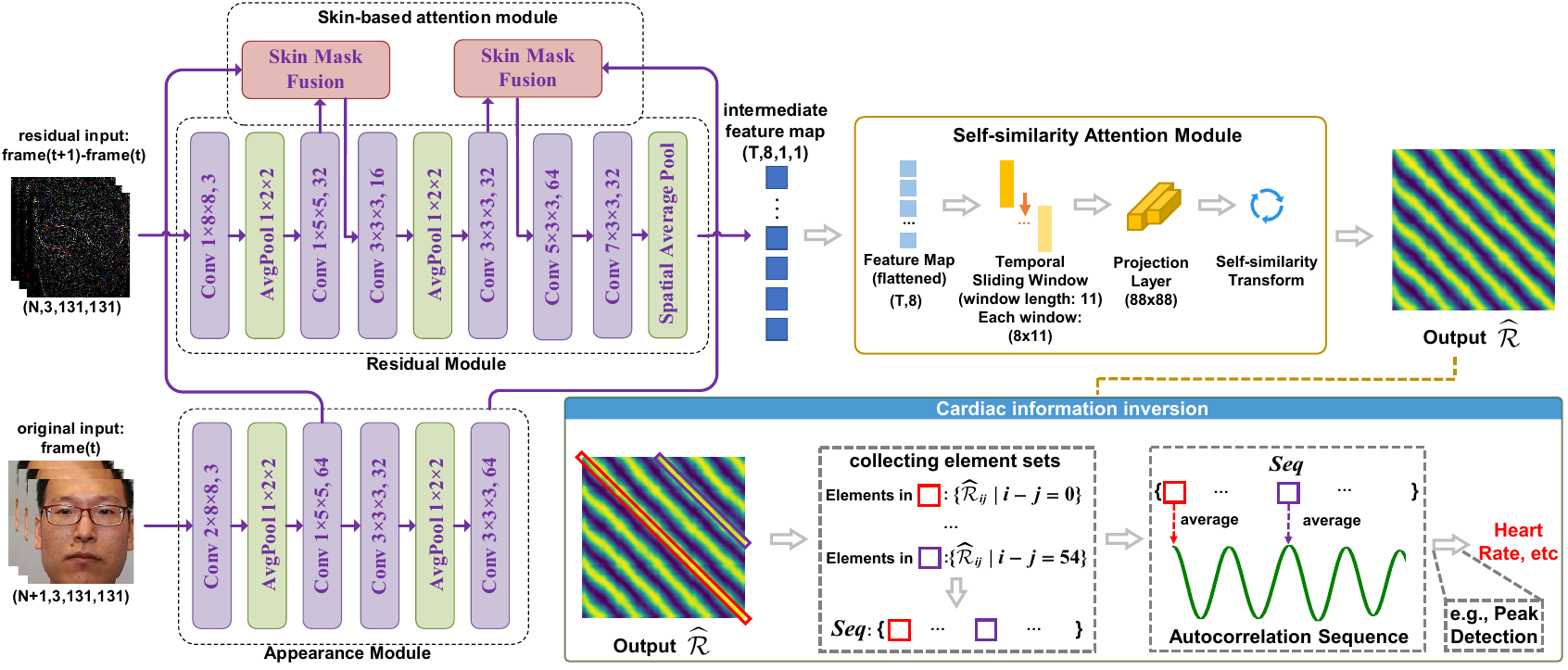}
    \vspace{-3mm}
    \caption{The adjusted baseline combined with proposed DOHA, and we detailed describe the process of ``cardiac information inversion'' (algorithm 1 in the main text) in the blue box.}
    \label{Appendix: Our adjustments}
    \vspace{-2mm}
\end{figure*}

In our experiments, the input size of the facial video is $[T \times 3 \times 131 \times 131]$, and we apply this size in all involved end-to-end rPPG networks (i.e., Deepphys, TS-CAN, PhysNet). This input size is different from those in the original settings (e.g., in Deepphys, the input size is $[T \times 3 \times 36 \times 36]$). In this section, we introduce our adjusted TS-CAN (the baseline in Tab. 1), as shown in Fig. \ref{Appendix: Original Baseline}. Moreover, the detailed implementation of our harmonious phase strategy (under inference) on this adjusted baseline is shown in Fig. \ref{Appendix: Our adjustments}.

\section{Algorithm for SSP map generation}
\label{Algorithm for SSP map generation}
In the main text, due to space limitation, we use Fig. \ref{SSP} to briefly introduce the generation of self-similarity physiological (SSP) map. In this section, we use pseudo code to specifically describe the generation progress, as shown in Algorithm. \ref{SCS Map}.

\begin{algorithm} 
	\caption{Self-similarity physiological map generation} 
	\label{SCS Map} 
	\begin{algorithmic}[1]
            \STATE {\bfseries Input:} Physiological signal $Sig: \{t_0, t_1, \cdots, t_{N-1}\}$, length of signal $N$, sliding window length $L_{w}$, similarity calculating function $sim(\cdot,\cdot)$
            \STATE Define output map size $L_o=N-{L_{win}}+1$, initialize slice list $List_{sl} = [0]_{L_o}$, initialize self-similarity physiological map $\mathcal{R}_{L_o \times L_o}=[0]_{L_o \times L_o}$
            \STATE $List_{sl}[i]$ $\leftarrow$ $\{ t_i, t_{i+1}, \cdots, t_{i+L_{win}-1} \}, \forall i \in [0, L_o-1]$
            \\ $\%$ self-similarity transform $\%$
            \FOR{$i \in [0, L_o-1]$}
            \FOR{$j \in [0, L_o-1]$}
            \STATE $\mathcal{R}_{ij} \leftarrow sim(List[i], List[j])$
            \ENDFOR
            \ENDFOR
            \STATE {\bfseries Output:} Self-similarity physiological map $\mathcal{R}_{L_o \times L_o}$
	\end{algorithmic}
\end{algorithm}

\section{Limitation of signal calibration}
\label{Limitation of signal calibration}
\begin{figure}[htbp]
    \centering
    \includegraphics[width=6.5cm]{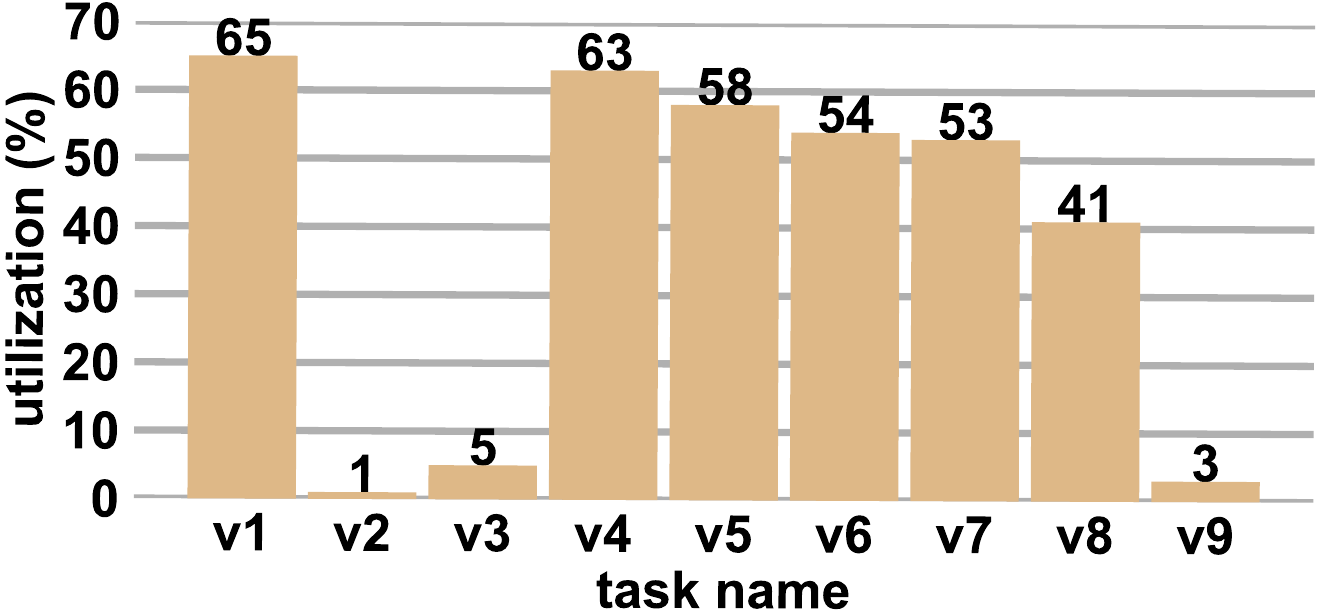}
    \vspace{-3mm}
    \caption{The utilization of VIPL-HR (source2) after calibration via GREEN. We can observe that few instances can be utilized in tasks v2, v3, v8, v9, which are all difficult tasks.}
    \label{Appendix: Data Utilization}
\end{figure}

The main text mentions that we can utilize traditional rPPG methods (e.g., GREEN) to calibrate the phase differentials among the ``ground truth'' and actual facial rPPG signals. However, this method is laborious and wasteful. The main problem is that such a standard for calibration is usually hard to acquire: the results in Tab. \ref{MainTable} indicate that these traditional methods cannot perform well under challenging scenarios such as instances in the VIPL-HR database. To verify this, we put the utilization of the database (VIPL-HR, source 2) after calibration, as shown in Fig. \ref{Appendix: Data Utilization}, from which we can observe that those challenging tasks (e.g., those in v2, v3) all get eliminated throughout the sifting. The decrement in data utilization could damage the diversity of training scenarios, thus impacting the model generalization.

\begin{figure}[htbp]
    \centering
    \includegraphics[width=7.85cm]{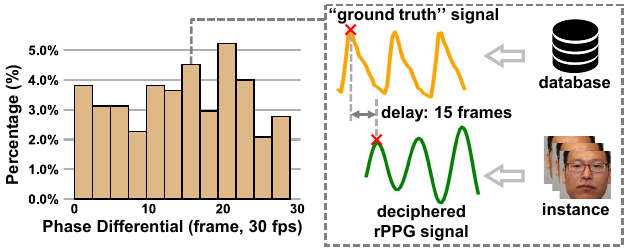}
    \vspace{-2mm}
    \caption{The probability density of phase delays among instances with VIPL database \cite{VIPL-HR} that can be deciphered (i.e., analyzed with high confidence) using GREEN \cite{GREEN}.}
    \label{Delay_Distribution}
    % \vspace{-2mm}
\end{figure}

\begin{table}[t]
\caption{Time consumption of DOHA, when training we use a batch size of 5, and video length is 75. The involved baseline is adjusted TS-CAN. The input frame size is 131 $\times$ 131.}
\vspace{-3mm}
\resizebox{0.9\columnwidth}{!}{
\begin{tabular}{c|cc}
\toprule[1.5pt]
                & \begin{tabular}[c]{@{}c@{}}Training time\\ per batch)\end{tabular} & \begin{tabular}[c]{@{}c@{}}Inference time \\ on 300 \\ frames video\end{tabular}  \\ \hline[1pt]
Baseline        & 0.526                                                 & 0.413                                                              \\  
DOHA+ (w/o HPS) & 1.764                                                                               & 0.413                                                                                             \\
DOHA+ (w/o GGH) & 1.935                                                                               & 0.607                                                                                             \\
DOHA+ (w/o IGH) & 0.960                                                                               & 0.607                                                                                             \\
DOHA+           & 1.934                                                                               & 0.607      \\ \bottomrule[1pt]                                                                        
\end{tabular}
}
% \vspace{-3mm}
\label{Timeconsumption}
\end{table}

Furthermore, we use the sifted data to present the delay distribution among the ``ground truth'' and actual facial rPPG signals, as shown in \ref{Delay_Distribution}. From the results, we can infer that such delay is genuinely uncertain and cannot be neglected (with the max delay about the whole cardiac period).

\vspace{-1mm}
\section{Time consumption of DOHA}
\label{Time consumption of DOHA}
The time consumption of DOHA is listed as Tab. \ref{Timeconsumption}. From the table, the primary time consumption is associated with IGH and GGH, as they involve instance-level gradient operations. However, these operations do not affect the time consumption during the inference stage. Conversely, DOHA-HPS also works during inference time but requires lower time consumption.

\balance
\end{document}